\documentclass{article}
\usepackage{KR26_authors_kit/kr}

\usepackage{times}
\usepackage{soul}
\usepackage{url}
\usepackage[hidelinks]{hyperref}
\usepackage[utf8]{inputenc}
\usepackage[small]{caption}
\usepackage{graphicx}
\usepackage{amsmath}
\usepackage{amsthm}
\usepackage{amssymb}
\usepackage{booktabs}
\usepackage{multirow}
\usepackage{algorithm}
\usepackage{algorithmic}
\usepackage{tikz}
\usepackage{microtype}
\usepackage{xcolor}

\usetikzlibrary{shapes,arrows,positioning,fit,calc}
\theoremstyle{definition}
\newtheorem{definition}{Definition} 
\newtheorem{example}{Example}
\newtheorem{theorem}{Theorem}
\newtheorem{proposition}{Proposition}

\title{Causal Discovery as Dialectical Aggregation: A Quantitative Argumentation Framework}

\author{%
Sheng Wei$^{1,2,4}$\and
Yulin Chen$^1$\and
Beishui Liao$^{2,3,4}$\thanks{The corresponding author is Beishui Liao (e-mail: baiseliao@zju.edu.cn).}\\
\affiliations
$^1$The College of Computer Science and Technology, Zhejiang University\\
$^2$ZLAIRE, Zhejiang University\\
$^3$School of Philosophy, Zhejiang University\\
$^4$The State Key Lab of Brain-Machine Intelligence\\
\emails
\{shengwei, yulinchen, baiseliao\}@zju.edu.cn
}

\begin{document}

\maketitle

\begin{abstract}
Constraint-based causal discovery is brittle in finite-sample regimes because erroneous conditional-independence (CI) decisions can cascade into substantial structural errors.
We propose Quantitative Argumentation for Causal Discovery (QACD), a semantics-driven framework that represents CI outcomes as graded, defeasible arguments rather than irreversible constraints.
QACD maps statistical test outcomes to argument strengths and aggregates conflicting evidence through connectivity-mediated witness propagation, producing a fixed-point acceptability labeling over candidate adjacencies.
Experiments on standard benchmark Bayesian networks suggest that QACD improves structural coherence and interventional reliability in several noisy or inconsistent CI regimes, while remaining competitive with classical constraint-based, hybrid, and prior argumentation-based baselines.
\end{abstract}

\section{Introduction}
\label{sec:introduction}

Causal discovery, the task of inferring causal structure from observational data, is a central problem in Artificial Intelligence, as it enables reasoning systems to move beyond statistical association toward interventional and counterfactual reasoning~\cite{neuberg2003causality,yang2025nonlinear}. 
The objective is to reconstruct the underlying Directed Acyclic Graph (DAG), or more practically its Markov equivalence class—typically represented as a Completed Partially Directed Acyclic Graph (CPDAG)—from independent and identically distributed (i.i.d.) samples drawn from an unknown joint distribution. 
Such causal models underpin applications ranging from gene regulatory network analysis~\cite{sachs2005causal} to economic and social policy evaluation~\cite{athey2017state}.

Two paradigms dominate the literature. 
Score-based methods search for a graph maximizing a penalized likelihood score~\cite{chickering2002optimal,zheng2018dags}, while constraint-based methods infer structure by testing conditional-independence (CI) relations~\cite{spirtes2000causation,colombo2014order}. 
Within the latter family, the Peter–Clark (PC) algorithm remains a canonical reference due to its efficiency and asymptotic guarantees: under standard assumptions, PC is sound and consistent when CI queries are answered by an oracle.

In practice, however, CI evidence obtained from statistical testing is inherently uncertain. 
Finite samples, measurement error, and mismatches between test assumptions and data routinely produce both false independences and false dependences. 
Constraint-based pipelines nevertheless treat accepted CI statements monotonically: once an edge is removed or a separating set is fixed, later evidence cannot overturn that commitment. 
Under finite samples, such monotonic reasoning is brittle, as early CI errors can accumulate into globally inconsistent commitments~\cite{ling2024causal} and cascade through subsequent orientation steps~\cite{ding2024causal,uhler2013geometry}.

We argue that this brittleness stems from a fundamental conceptual mismatch: uncertain statistical inferences are treated as hard logical constraints rather than defeasible evidence ~\cite{uhler2013geometry}. Consequently, causal discovery should be reframed as a problem of defeasible reasoning rather than pure constraint satisfaction.

This motivates an argumentation-theoretic perspective, in which CI outcomes are represented as graded pieces of evidence whose conflicts must be resolved dialectically~\cite{dung1995acceptability,pollock1987defeasible,wang2025fuzzy}. 
Existing uses of argumentation in causal discovery, however, have mostly focused on later stages such as edge orientation or on qualitative rather than quantitative evidence ~\cite{russoargumentative}. 
What therefore remains largely unexplored is how to integrate graded statistical evidence directly into the skeleton discovery stage through an explicit argumentation semantics.

To address this gap, we propose \emph{Quantitative Argumentation for Causal Discovery} (QACD), which evaluates candidate adjacencies by dialectically aggregating uncertain CI evidence. Its main contributions are:
\begin{itemize}
    \item \textbf{A quantitative argumentation formalization of CI-based causal discovery.} 
    CI outcomes are interpreted as graded, defeasible arguments rather than irreversible constraints, enabling reasoning under inconsistent statistical evidence.
    
    \item \textbf{Connectivity-mediated propagated attenuation.} 
    We introduce a structure-dependent conflict mechanism through which global graph context modulates the strength of local CI claims.
    
    \item \textbf{A fixed-point dialectical semantics for skeleton revision.} 
    Edge acceptability is revised through finite-horizon or convergent updates, yielding a stable skeleton that reflects aggregated evidential conflicts rather than greedy deletions.
\end{itemize}

The remainder of this paper is organised as follows.
Section~\ref{sec:preliminaries} reviews causal graphs and background on quantitative argumentation.
Section~\ref{sec:methodology} presents QACD and its dialectical semantics.
Section~\ref{sec:experiments} evaluates the approach empirically, followed by related work and conclusions.

\section{Preliminaries}
\label{sec:preliminaries}

This section reviews the causal-graphical and argumentation-theoretic notions used throughout the paper and fixes notations.

\subsection{Causal Graphs}
We assume that the underlying causal structure over observed variables can be represented by a DAG~\cite{pearl1988probabilistic}. Let $G=(\mathbf{V},\mathbf{E})$ be a DAG, where $\mathbf{V}=\{V_1,\dots,V_n\}$ is a set of random variables and $\mathbf{E}\subseteq \mathbf{V}\times\mathbf{V}$ is a set of directed edges. An edge $X\to Y$ indicates that $X$ is a (direct) cause of $Y$ in $G$.

For a node $X\in\mathbf{V}$, a (possibly non-simple) path between two nodes is a sequence of adjacent nodes, where edge directions are ignored. A node $M$ on a path is a \emph{collider} on that path if the two incident edges are oriented $X\to M \leftarrow Y$; otherwise $M$ is a \emph{non-collider}.

\begin{definition}[d-separation]
\label{def:d-separation}
Let $X,Y\in\mathbf{V}$ and $\mathbf{Z}\subseteq \mathbf{V}\setminus\{X,Y\}$. A path $p$ between $X$ and $Y$ is \emph{active} (or \emph{d-connecting}) given $\mathbf{Z}$ if:
\begin{enumerate}
    \item every non-collider on $p$ is not in $\mathbf{Z}$; and
    \item every collider on $p$ is in $\mathbf{Z}$ or has a descendant in $\mathbf{Z}$.
\end{enumerate}
If there exists no active path between $X$ and $Y$ given $\mathbf{Z}$, then $X$ and $Y$ are \emph{d-separated} by $\mathbf{Z}$ in $G$, denoted $X \perp\!\!\!\perp_G Y \mid \mathbf{Z}$.
\end{definition}

Graphical separation is linked to probabilistic conditional independence through standard assumptions.

\begin{definition}[Causal Markov Assumption]
\label{def:causal_markov}
Let $P$ be the joint probability distribution over $\mathbf{V}$ and $G$ be a causal DAG. The distribution $P$ satisfies the \emph{Causal Markov Assumption} with respect to $G$ if every d-separation relation in $G$ implies a conditional independence in $P$. That is, for all disjoint sets of variables $X,Y\in\mathbf{V}$ and $\mathbf{Z}\subseteq \mathbf{V}\setminus\{X,Y\}$:
\begin{equation}
    X \perp\!\!\!\perp_G Y \mid \mathbf{Z} \implies X \perp\!\!\!\perp_P Y \mid \mathbf{Z}.
\end{equation}
\end{definition}

While the Markov assumption allows us to read independencies from the graph, we also require the converse to infer the graph structure from observed independencies.

\begin{definition}[Faithfulness Assumption]
\label{def:faithfulness}
The distribution $P$ is said to be \emph{faithful} to the DAG $G$ if every conditional independence relation in $P$ implies a d-separation in $G$. That is, for all distinct variables $X,Y\in V$ and all $Z\subseteq V\setminus\{X,Y\}$:
\begin{equation}
    X \perp\!\!\!\perp_P Y \mid \mathbf{Z} \implies X \perp\!\!\!\perp_G Y \mid \mathbf{Z}.
\end{equation}
\end{definition}
While we assume faithfulness as a background idealization, our framework is explicitly designed to operate under finite-sample regimes where CI test outcomes may violate faithfulness due to statistical error.

\begin{definition}[Markov Equivalence]
\label{def:markov_equivalence}
Two DAGs $G_1$ and $G_2$ over $\mathbf{V}$ are \emph{Markov equivalent} if they entail the same set of d-separation relations. A necessary and sufficient condition for Markov equivalence is that $G_1$ and $G_2$ possess the same \emph{skeleton} (adjacency structure ignoring direction) and the same set of \emph{v-structures} (unshielded colliders of the form $X \to Z \leftarrow Y$ where $X$ and $Y$ are not adjacent). The set of all DAGs Markov equivalent to a graph $G$ constitutes its \emph{Markov Equivalence Class} (MEC), denoted by $[G]$.
\end{definition}

Since observational data (under the Markov and Faithfulness assumptions) only allows us to identify the MEC rather than a unique DAG, we utilise a graphical object to represent the entire class.

\begin{definition}[Completed Partially Directed Acyclic Graph]
\label{def:cpdag}
A Markov equivalence class $[G]$ can be uniquely represented by a CPDAG $G^\ast$, which has the same skeleton as any DAG in $[G]$, and whose edges are directed if and only if they are compelled across the class.
\end{definition}

Our goal is to recover the target CPDAG $G^\ast$ from finite samples~\cite{anand2025causal}. Constraint-based methods use CI tests to remove adjacencies and separating sets to orient edges; under finite samples, CI errors motivate the non-monotonic treatment developed in the next section.

\textbf{Orientation Rules (Meek).}
Given an undirected skeleton and a collection of separating sets, standard constraint-based causal discovery procedures apply a set of sound graphical orientation rules, known as \emph{Meek rules}~\cite{meek1995causal}, to orient all compelled edges and obtain the unique CPDAG representing the corresponding MEC. These rules are deterministic, purely graphical, and do not introduce
additional statistical assumptions.

\textbf{Conditional Independence Tests and \emph{p}-values.}
Throughout this paper, CI relations are assessed using standard statistical tests, each returning a \emph{p}-value. We interpret the \emph{p}-value not as a hard decision criterion, but as a graded indicator of evidential strength for a CI claim: larger \emph{p}-values correspond to stronger statistical support for conditional independence~\cite{jabbari2017discovery}.

\subsection{Quantitative Argumentation Frameworks} 
To situate our method more clearly within formal argumentation, we briefly recall the attack-only quantitative setting used in this paper. 
We require only a minimal set of notions: arguments, attacks, weights, and gradual acceptability.

\begin{definition}[Quantitative Argumentation Framework] 
In this paper, we use a lightweight attack-only quantitative argumentation framework (QAF) of the form
\[
\mathcal{Q}=\langle A,R,w,\rho\rangle,
\]
where \(A\) is a finite set of arguments, \(R\subseteq A\times A\) is an attack relation, \(w:A\to[0,1]\) assigns intrinsic strengths to arguments, and \(\rho:R\to[0,1]\) assigns strengths to attacks.
If \((a,b)\in R\), then \(a\) attacks \(b\) with strength \(\rho(a,b)\). 
\end{definition} 

\begin{definition}[Gradual Semantics] 
Let \(\mathcal{Q}=\langle A,R,w,\rho\rangle\) be a QAF. 
A gradual semantics assigns to each argument \(a\in A\) an acceptability degree \(s(a)\in[0,1]\), rather than a binary status such as accepted or rejected. 
Such a semantics may be specified by an operator \(\Gamma:[0,1]^{|A|}\to[0,1]^{|A|}\), whose iterates determine the acceptability degrees of arguments under the influence of weighted attacks. 
Intuitively, an argument remains acceptable to the extent that it withstands the strength of its attackers. 
\end{definition} 

The framework proposed in this paper instantiates this attack-only and gradual setting for causal discovery. 
Section~\ref{sec:methodology} treats CI statements as graded attackers of adjacency hypotheses and defines a state-dependent acceptability update over candidate edges. 
In particular, the effective attack strengths in QACD may depend on the current graph state through witness-based propagation.

\begin{figure*}[t]
\centering
\small
\begin{tikzpicture}[
    font=\small,
    box/.style={
        rectangle, draw, rounded corners,
        align=center, inner sep=4pt, fill=blue!8,
        minimum width=33mm, minimum height=11mm
    },
    arrow/.style={->,>=latex, line width=0.5pt},
    dashedarrow/.style={->,>=latex, dashed, line width=0.5pt},
    group/.style={
        rectangle, draw, rounded corners,
        inner sep=6pt, dashed
    },
    node distance=10mm and 12mm
]

\node[box] (data) {Data\\$\mathcal{D}$};
\node[box, right=of data] (citest) {CI tests\\$\Rightarrow$ arguments $\tau$\\strength $\sigma_0(\tau)$};
\node[box, right=of citest] (cand) {Phase I:\\Candidate graph\\$G_{\mathrm{cand}}$};

\node[box, below=14mm of citest] (init) {Initialize\\$\mathbf{S}^{(0)}$ from\\$G_{\mathrm{cand}}$};
\node[box, left=of init] (iter) {Phase II:\\Dialectical update\\$\mathbf{S}^{(t+1)}=\mathbf{S}^{(t)}\odot\mathbf{A}(\mathbf{S}^{(t)})$\\
\footnotesize direct attacks + witness propagation};
\node[box, right=of init] (out) {Threshold $\delta_0$\\$\Rightarrow$ skeleton\\v-structures + Meek rules\\$\Rightarrow$ CPDAG};

\draw[arrow] (data) -- (citest);
\draw[arrow] (citest) -- (cand);
\draw[arrow] (cand) -- (init);
\draw[arrow] (init) -- (iter);

\draw[dashedarrow] (iter.north) to[bend left=20] node[above, font=\footnotesize]{iterate} (iter.north east);

\node[group, fit=(data) (citest)] (g0) {};
\node[above right=-1mm and -1mm of g0.north west, anchor=south west, font=\footnotesize]
{Evidence Construction};

\node[group, fit=(cand)] (g1) {};
\node[above right=-1mm and -1mm of g1.north west, anchor=south west, font=\footnotesize]
{Candidate Generation};

\node[group, fit=(iter) (init)] (g2) {};
\node[above right=-1mm and -1mm of g2.north west, anchor=south west, font=\footnotesize]
{Dialectical Aggregation};

\node[group, fit=(out)] (g3) {};
\node[above right=-1mm and -1mm of g3.north west, anchor=south west, font=\footnotesize]
{Graph Recovery};

\draw[arrow] (g2.east) -- (g3.west);

\end{tikzpicture}
\caption{Overview of QACD. Phase~I builds a permissive candidate graph. Phase~II aggregates CI evidence via quantitative argumentation with direct attacks and witness-based propagation, iteratively updating edge acceptability before thresholding and orientation.}
\label{fig:qacd_pipeline}
\end{figure*}
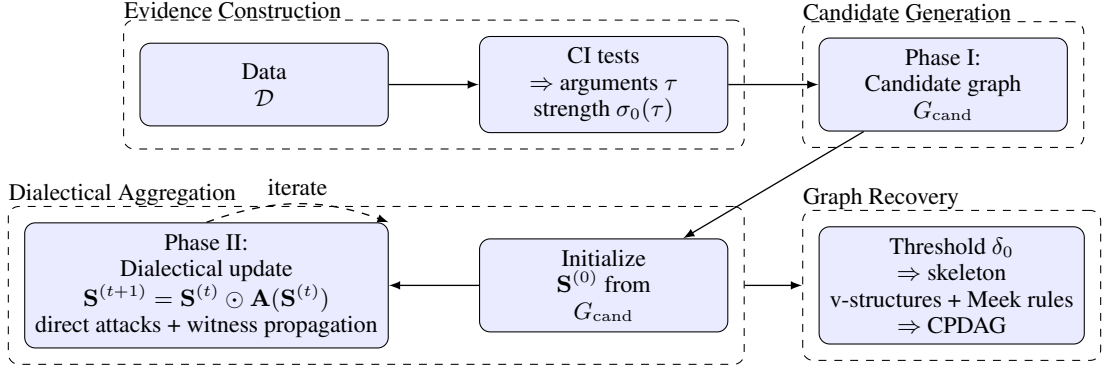

\section{Methodology}
\label{sec:methodology}
We now present QACD, which learns causal skeletons by dialectically aggregating uncertain CI evidence.
As shown in Figure~\ref{fig:qacd_pipeline}, Phase~I constructs a permissive candidate graph and collects CI evidence, while Phase~II iteratively revises edge acceptability under direct and witness-propagated attacks before thresholding and orientation.
Algorithm~\ref{alg:qacd} implements the QACD semantics.

\subsection{Problem Setup and Notation}
\label{sec:problem_setup}

Let $\mathcal{V}$ be a finite set of observed variables generated by an unknown causal DAG $\mathcal{G}^\ast$ satisfying the Causal Markov and Faithfulness assumptions.
Given i.i.d.\ samples, a CI testing procedure returns a statistic (e.g., a $p$-value) for statements of the form $X \perp\!\!\!\perp Y \mid \mathbf{Z}$.

Our goal is to recover the MEC of $\mathcal{G}^\ast$, represented as a CPDAG, while avoiding brittle monotonic decisions caused by finite-sample CI errors.

\paragraph{Argumentative Representation.}
We write $X{-}Y$ for an undirected adjacency and $X \to Y$ for a directed edge.
QACD maintains a symmetric \emph{acceptability matrix}
$\mathbf{S} \in [0,1]^{|\mathcal{V}|\times|\mathcal{V}|}$,
where $S_{XY}$ denotes how acceptable the hypothesis “$X$ is adjacent to $Y$” currently is.
Values close to $1$ indicate strong support; values near $0$ indicate rejection.
We refer to $S_{XY}$ simply as the acceptability of the edge hypothesis $e_{XY}$.

QACD can be viewed as a two-sorted quantitative argumentative system whose state space consists of edge hypotheses and representative CI-derived independence arguments. Edge hypotheses $e_{XY}$ are the claims whose acceptability is iteratively updated, whereas CI arguments $\tau_{XY}$ carry fixed initial strengths $\sigma_0(\tau_{XY})$ and act as graded attackers. Witness-based propagation is not introduced as a separate argument node, but as a state-dependent operator that modulates how CI-derived attack pressure extends to adjacent edge hypotheses through the current graph structure.
We distinguish three classes of argumentative objects:

\begin{enumerate}
    \item \textbf{Edge Hypotheses ($e_{XY}$):} The target claims to be justified or defeated.
    
    \item \textbf{Independence Arguments ($\tau$):} Statistical findings that attack edge hypotheses. An independence argument $\tau_{XY}$ is a tuple defined by a conditioning set $\mathbf{Z}_\tau \subseteq \mathcal{V}\setminus\{X,Y\}$ and a test outcome asserting $X \perp\!\!\!\perp Y \mid \mathbf{Z}_\tau$. We assign each $\tau_{XY}$ a \emph{base strength} $\sigma_0(\tau_{XY}) \in [0,1]$ derived from its test statistic, treating it as a defeasible reason to reject $e_{XY}$.
    
    \item \textbf{Structural Modifiers:} Context-dependent modifiers induced by the evolving graph structure, specifically by witness paths. These modifiers do not constitute separate argument nodes; instead, they modulate how CI-derived attack pressure is propagated to structurally implicated edge hypotheses when independence evidence conflicts with strong local connectivity.
\end{enumerate}

In the subsequent methodology, we detail how specific arguments are instantiated (Phase~I) and dialectically aggregated (Phase~II).

\subsection{Phase I: Candidate Graph Generation}
\label{subsec:candidate_generation}

Phase~I constructs a permissive candidate skeleton to avoid premature deletions.

For each pair $\{X,Y\}$, we compute
\begin{equation}
p_{X Y}^{\min }=\min _{Z \subseteq V \backslash\{X, Y\},|Z| \leq K} p(X \perp\!\!\!\perp Y \mid Z),
\label{eq:p_min}
\end{equation}
where $K$ is the maximum conditioning-set size used in Phase I, and include $X{-}Y$ in the candidate graph $G_{\text{cand}}$ whenever
$p^{\min}_{XY} < \delta_{\text{cand}}$ (candidate threshold).
This rule is intentionally conservative: if any conditioning set suggests dependence, the edge is retained as a viable hypothesis.

Each variable pair additionally yields a single representative CI statement $\tau_{XY}$, corresponding to the conditioning set that provides the strongest empirical support for independence, as quantified by the largest associated $p$-value. In Phase~II, $\tau_{XY}$ is treated as an \emph{independence argument}, whose evidential strength is determined by the value of its corresponding test statistic.

For each unordered pair \((X,Y)\), QACD retains a single representative CI argument, namely the statement associated with the largest observed p-value among the tested conditioning sets.
Accordingly, the global CI pool is denoted by \(\mathcal{T}=\{\tau_{XY}:X<Y\}\).
Thus, the update uses one representative independence argument per unordered pair.

\subsection{Phase II: Dialectical Aggregation via Quantitative Argumentation}
\label{subsec:dialectical_aggregation}

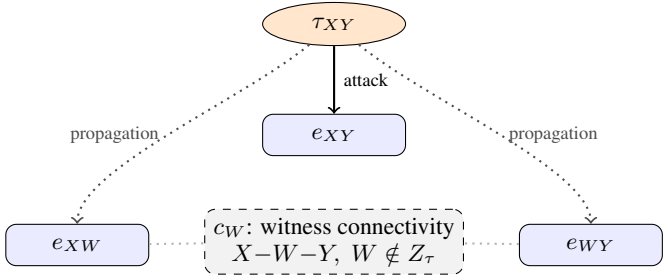
\begin{figure}[t]
\centering
\begin{tikzpicture}[
    node distance=0.9cm,
    every node/.style={font=\small},
    claim/.style={
        rectangle, draw,
        rounded corners,
        minimum width=1.9cm,
        minimum height=0.55cm,
        fill=blue!8,
        align=center
    },
    argument/.style={
        ellipse, draw,
        minimum width=1.9cm,
        minimum height=0.55cm,
        fill=orange!20,
        align=center
    },
    context/.style={
        rectangle, draw,
        dashed,
        rounded corners,
        minimum width=2.8cm,
        minimum height=0.7cm,
        fill=gray!10,
        align=center
    },
    attack/.style={->, thick},
    undercut/.style={->, thick, dashed, color=gray!70},
    propagate/.style={->, thick, dotted, color=black!70},
    support/.style={-, dotted, thick, color=gray!60}
]

\node[claim] (eXY) {$e_{XY}$};
\node[argument, above=of eXY] (tauXY) {$\tau_{XY}$};

\node[claim, below left=0.9cm and 1.5cm of eXY] (eXW) {$e_{XW}$};
\node[claim, below right=0.9cm and 1.5cm of eXY] (eWY) {$e_{WY}$};

\node[context, below=0.7cm of eXY] (cW)
{$c_W$: witness connectivity\\$X{-}W{-}Y,\; W\notin Z_\tau$};

\draw[attack] (tauXY) -- node[right, font=\scriptsize] {attack} (eXY);

\draw[support] (eXW) -- (cW);
\draw[support] (eWY) -- (cW);

\draw[propagate]
(tauXY) .. controls +(-1.2,-1.0) and +(0,1.0) ..
node[left, font=\scriptsize] {propagation} (eXW);

\draw[propagate]
(tauXY) .. controls +(1.2,-1.0) and +(0,1.0) ..
node[right, font=\scriptsize] {propagation} (eWY);

\end{tikzpicture}
\caption{A representative CI argument $\tau_{XY}$ directly attenuates the target edge $e_{XY}$ and, when a witness connection $X\!-\!W\!-\!Y$ with $W \notin Z_{\tau}$ is present, also propagates attenuation to the bridging edges $e_{XW}$ and $e_{WY}$ in proportion to the witness connectivity $c_W$.}
\label{fig:argumentation_example}
\end{figure}

Having constructed the candidate skeleton, we now resolve conflicts among edges by instantiating a state-dependent QAF. In this framework, the set of arguments is fixed, but the effective attack relations depend on the evolving acceptability state of the graph. Consequently, the influence of evidence is inherently context-sensitive: the same CI statement may have different effects depending on the surrounding structure.

Phase II uses a gradual semantics in which edge hypotheses are revised under weighted CI attacks and context-dependent witness-based propagated attenuation.

Figure~\ref{fig:argumentation_example} illustrates this interaction between adjacency hypotheses, CI-based rebuttals, and witness-induced propagated attenuation.

\begin{example}[Running example]
Suppose the current candidate graph contains the triangle $X\!-\!Y$, $X\!-\!W$, and $W\!-\!Y$. Assume that Phase~I retains a representative CI statement
$\tau_{XY} : X \perp\!\!\!\perp Y \mid Z_{\tau}$
with a large $p$-value, and that $W \notin Z_{\tau}$. Then $\tau_{XY}$ provides strong direct pressure against the target edge hypothesis $e_{XY}$. If the current acceptability scores of $e_{XW}$ and $e_{WY}$ are still high, $W$ becomes an eligible witness, so part of the same pressure is propagated to the two bridging edges. In this way, QACD does not only revise the disputed edge $X\!-\!Y$ in isolation; it also weakens the local configuration that sustains the conflicting pattern.
\end{example}

Let $\tau_{XY}$ denote the representative CI statement retained for $(X,Y)$, with base strength $\sigma_0(\tau_{XY})\in[0,1]$.
This representative independence statement is treated as an attack argument against $e_{XY}$, weakening its acceptability.
Beyond this direct attack, QACD introduces witness-based propagated attenuation (formalized in Definition~\ref{def:undercut}), which allows strong independence evidence to weaken structurally implicated bridging edges under inconsistent CI evidence.

Formally, the graded semantics is computed via an iterative dialectical update:
\begin{equation}
\mathbf{S}^{(t+1)} \;=\; \mathbf{S}^{(t)} \odot \mathbf{A}\!\left(\mathbf{S}^{(t)}\right),
\label{eq:dialectical_update}
\end{equation}
where $\odot$ is elementwise multiplication, $\mathbf{S}^{(t)}$ denotes the current acceptability and $\mathbf{A}(\mathbf{S}^{(t)}) \in [0,1]^{|\mathcal{V}|\times|\mathcal{V}|}$ is an attenuation matrix determined by (i) direct attacks induced by independence arguments and (ii) connectivity-mediated propagated attenuation triggered by short witness connections.
We stop after $T_{\max}$ iterations or when $\|\mathbf{S}^{(t+1)}-\mathbf{S}^{(t)}\|_\infty$ is below a small tolerance.
That is, an edge hypothesis retains its current acceptability unless it is weakened by an applicable defeasible influence.
The iteration continues until the acceptability state stabilizes up to the prescribed stopping criterion.

We now describe the quantitative argumentation semantics used to revise $\mathbf{S}$ in the presence of conflicting CI evidence.

For any fixed acceptability state \(S\in[0,1]^{|V|\times |V|}\), QACD induces a state-dependent QAF
\[
\mathcal{Q}(S)=\langle \mathcal{A},R_S,w,\rho_S\rangle,
\]
where \(\mathcal{A}=\mathcal{E}\cup\mathcal{T}\), \(\mathcal{E}=\{e_{XY}:X<Y\}\), and \(\mathcal{T}=\{\tau_{XY}:X<Y\}\).
Here \(e_{XY}\) is the edge-hypothesis argument for the adjacency claim between \(X\) and \(Y\), and \(\tau_{XY}\) is the representative CI-derived independence argument for the same unordered pair retained in Phase~I.
The weight function is defined by \(w(e_{XY})=S^{(0)}_{XY}\) and \(w(\tau_{XY})=\sigma_0(\tau_{XY})\).

The attack relation \(R_S\) contains each direct attack \((\tau_{XY},e_{XY})\).
Moreover, whenever \(W\notin Z_{\tau}\) is an eligible witness for \((X,Y)\) at state \(S\), \(R_S\) also contains the propagated attacks \((\tau_{XY},e_{XW})\) and \((\tau_{XY},e_{WY})\).
The corresponding effective attack strengths are
\[
\rho_S(\tau_{XY},e_{XY})=\lambda\sigma_0(\tau_{XY})
\]
for direct attacks, and
\[
\rho_S(\tau_{XY},e_{XW})
=
\rho_S(\tau_{XY},e_{WY})
=
\frac{\lambda}{2}\sigma_0(\tau_{XY})S_{XW}S_{WY}
\]
for propagated attacks.

\paragraph{Independence arguments and direct attacks.}
Each independence argument $\tau_{XY}$ provides a weighted direct attack on the edge hypothesis $e_{XY}$.
In the current implementation, its CI-test $p$-value is mapped to a base strength
\begin{equation}
\sigma_{0}\left(\tau_{X Y}\right)=\left\{\begin{array}{ll}
0, & p\left(\tau_{X Y}\right)<\alpha, \\
\frac{p\left(\tau_{X Y}\right)-\alpha}{1-\alpha}, & p\left(\tau_{X Y}\right) \geq \alpha,
\end{array}\right.
\label{eq:tauxy}
\end{equation}
so that only non-rejected CI claims contribute positive independence strength, and among them larger $p$-values yield stronger defeasible support for independence.

This strength is interpreted as a direct rebuttal on the edge hypothesis $e_{XY}$.
Since the current implementation retains a single representative CI argument for each unordered pair $(X,Y)$, the direct attenuation term is
\begin{equation}
A^{\text{dir}}_{XY}
= 1-\lambda \sigma_0(\tau_{XY}).
\label{eq:direct_attack}
\end{equation}

The parameter $\lambda$ controls how aggressively defeasible evidence propagates through the structure.
Importantly, this aggregation does not represent a vote or a probability, but a cumulative weakening of justification.
This form yields a bounded, order-independent attenuation semantics.
When multiple applicable direct or propagated factors affect the same edge hypothesis, they combine multiplicatively to weaken its acceptability.

\textbf{Aggregation rationale.}
Eq.~\eqref{eq:direct_attack} implements a bounded, proportional attenuation of the edge hypothesis by its representative CI argument. This form preserves acceptability within $[0,1]$ and composes naturally with the iterative update in Eq.~\eqref{eq:dialectical_update}.

In argumentation theory, a rebuttal directly attacks a claim, whereas an undercut challenges the applicability of such an attack without asserting the opposite conclusion~\cite{amgoud2000modelling}. In the causal discovery setting, strong CI evidence may therefore undercut not the edge hypothesis itself, but the structural relations that sustain conflicting connectivity patterns.

\begin{definition}[Witness-Based Propagated Attenuation via Witness Connectivity]
\label{def:undercut}
Fix an independence argument $\tau_{XY}$ with conditioning set $\mathbf{Z}_\tau$
and base strength $\sigma_0(\tau_{XY})$.
At iteration $t$, a node $W\in\mathcal{V}\setminus\{X,Y\}$ is an \emph{eligible witness} for $(X,Y)$ with respect to $\tau$ if
(i) it forms a length-2 connection $X{-}W{-}Y$ in the current candidate structure (equivalently, $S^{(t)}_{XW}>0$ and $S^{(t)}_{WY}>0$), and
(ii) $W \notin \mathbf{Z}_\tau$.
Such a witness induces a \emph{connectivity support} for $X$--$Y$ measured by
\begin{equation}
u^{(t)}_{\tau}(W) \;=\; S^{(t)}_{XW}\cdot S^{(t)}_{WY}.
\label{eq:witness_strength}
\end{equation}
\end{definition}
Intuitively, when a strong witness connection $X{-}W{-}Y$ remains acceptable, the CI argument $\tau_{XY}$ is not interpreted in isolation. Instead, its influence is allowed to propagate to the bridging edge hypotheses $e_{XW}$ and $e_{WY}$ in proportion to both the argument strength and the current witness connectivity. This state-dependent propagation weakens the local structural support underlying the competing witness pattern, as illustrated in Figure~\ref{fig:argumentation_example}.

\textbf{Design Rationale for Witness-Based Propagated Attenuation.}
Restricting witnesses to length-2 connections is a principled and computationally grounded design choice.
Such paths are the minimal non-trivial connectivity units in causal graphs and align with the local reasoning patterns exploited by constraint-based methods, where CI errors can cascade through short neighborhoods.
When strong independence evidence conflicts with a strongly supported length-2 connectivity pattern outside its conditioning set, the inconsistency can often be resolved by weakening one of the bridging relations, thereby restoring local coherence under inconsistent CI evidence.
Longer paths can be viewed as compositions of length-2 units and are therefore indirectly captured through iterative propagation; extending propagation to longer paths is possible but computationally more expensive.

\begin{algorithm}[t]
\caption{QACD: Quantitative Argumentation for Causal Discovery}
\label{alg:qacd}
\begin{algorithmic}[1]
\STATE \textbf{Input:} data $\mathcal{D}$ over variables $\mathcal{V}$; CI test procedure; max conditioning size $K$; candidate threshold $\delta_{\text{cand}}$; final threshold $\delta_0$; aggressiveness $\lambda$; max iterations $T_{\max}$; tolerance $\varepsilon$
\STATE \textbf{Output:} estimated CPDAG $\hat{G}^\ast$

\STATE \textbf{Phase I: Candidate Generation}
\STATE Initialize $G_{\text{cand}} \leftarrow$ empty graph
\STATE Initialize CI-statement pool $\mathcal{T}\leftarrow \emptyset$
\FORALL{unordered pairs $\{X,Y\}\subseteq \mathcal{V}$}
    \STATE Compute $p_{X Y}^{\min }$ using Eq.~\eqref{eq:p_min}
    \STATE Let $\mathbf{Z}^\star_{XY}\in \arg\max p(X \perp\!\!\!\perp Y \mid \mathbf{Z})$ and add $\tau_{XY}$ to $\mathcal{T}$
    \IF{$p^{\min}_{XY} < \delta_{\text{cand}}$}
        \STATE Add undirected edge $X{-}Y$ to $G_{\text{cand}}$
    \ENDIF
\ENDFOR

\STATE Initialize acceptability matrix $\mathbf{S}^{(0)}$ from $G_{\text{cand}}$

\STATE \textbf{Phase II: Dialectical Aggregation}
\FOR{$t=0$ \TO $T_{\max}-1$}
    \STATE Initialize attenuation matrix $\mathbf{A}^{(t)} \leftarrow \mathbf{1}$
    \FORALL{CI statements $\tau_{XY}$ in the pool $\mathcal{T}$}
        \STATE Compute base strength $\sigma_0(\tau_{XY})$ using Eq.~\eqref{eq:tauxy}
        \STATE Update direct attenuation on $(X,Y)$ using Eq.~\eqref{eq:direct_attack}
        \FORALL{witness nodes $W$ such that $X{-}W{-}Y$ in $G_{\text{cand}}$ \textbf{and} $W \notin \mathbf{Z}_\tau$}
            \STATE Update propagated attenuation on $(X,W)$ and $(W,Y)$ using Eq.~\eqref{eq:prop_attack}
        \ENDFOR
    \ENDFOR
    \STATE Update acceptability: $\mathbf{S}^{(t+1)} \leftarrow \mathbf{S}^{(t)} \odot \mathbf{A}^{(t)}$
    \IF{$\|\mathbf{S}^{(t+1)}-\mathbf{S}^{(t)}\|_\infty < \varepsilon$}
    \STATE \textbf{break}
    \ENDIF
\ENDFOR

\STATE \textbf{Skeleton and CPDAG Recovery}
\STATE Threshold $\mathbf{S}^{(T)}$ at $\delta_0$ to obtain skeleton $\hat{G}_{\text{skel}}$
\STATE Apply standard orientation rules to obtain CPDAG $\hat{G}^\ast$
\RETURN $\hat{G}^\ast$
\end{algorithmic}
\end{algorithm}

Although inspired by context-sensitive undercut-style reasoning, witness connectivity is used here as a state-dependent structural modifier that propagates CI-derived attack pressure to adjacent bridging edges.
In the present formulation, witness connectivity serves as a state-dependent propagation signal: when a strong witness pattern remains acceptable, part of the influence of the CI-derived attack on $e_{XY}$ is redistributed to the adjacent bridging edges.

\textbf{Propagation operator.}
The strength of a witness $W$ under the current acceptability is $u^{(t)}_{\tau}(W)$.
The propagated attenuation applied to each bridging edge induced by $\tau_{XY}$ and witness $W$ is defined as
\begin{align}
A_{X W \leftarrow \tau_{X Y}}^{\text {prop }}(W)=&A_{W Y \leftarrow \tau_{X Y}}^{\text {prop }}(W) \nonumber\\
=&1-\frac{\lambda}{2} \sigma_{0}\left(\tau_{X Y}\right) u_{\tau}^{(t)}(W),
\label{eq:prop_attack}
\end{align}
where the factor $\frac{1}{2}$ is a symmetric normalization choice that splits the propagated attenuation equally across the two bridging edges.
It is a simple heuristic design choice that preserves symmetry between $e_{XW}$ and $e_{WY}$.
We set $\lambda\in[0,1]$ to ensure $A^{\text{prop}}_{e \leftarrow \tau}(W)\in[0,1]$.
Direct and propagated attenuation factors are combined multiplicatively in the attenuation matrix $\mathbf{A}(\mathbf{S}^{(t)})$; if no factor applies to an edge, its attenuation is set to $1$.

For clarity, the overall attenuation applied to an edge hypothesis $e$ at iteration $t$ is the product of all direct and propagated factors that apply to that edge:
\[
A_e(S^{(t)})=\prod_{\tau\in\mathcal{T}_{\mathrm{dir}}(e)} A^{\mathrm{dir}}_{e\leftarrow\tau}\cdot \prod_{(\tau,W)\in\mathcal{T}_{\mathrm{prop}}(e)} A^{\mathrm{prop}}_{e\leftarrow\tau}(W),
\]
with the convention that an empty product equals $1$. Here, $\mathcal{T}_{\mathrm{dir}}(e)$ denotes the set of representative CI arguments that directly attenuate $e$, and $\mathcal{T}_{\mathrm{prop}}(e)$ denotes the set of witness-indexed propagated attenuation terms that apply to $e$ at iteration $t$.
Equation~(\ref{eq:dialectical_update}) is therefore understood entrywise as $S^{(t+1)}_e=S^{(t)}_e\cdot A_e(S^{(t)})$ for each edge hypothesis $e$.

Basic properties of the iterative semantics follow directly from the form of the attenuation factors.

\begin{theorem}[Boundedness]
\label{thm:bounded}
If $\mathbf{S}^{(0)}\in[0,1]^{|\mathcal{V}|\times|\mathcal{V}|}$, then $\mathbf{S}^{(t)}\in[0,1]^{|\mathcal{V}|\times|\mathcal{V}|}$ for all $t\ge 0$.
\end{theorem}
\begin{proof}
By construction, $\mathbf{A}(\mathbf{S}^{(t)})$ is obtained by multiplying factors of the form $(1-\lambda\sigma_0(\tau))$ and $\bigl(1-\frac{\lambda}{2}\sigma_0(\tau)u^{(t)}_{\tau}(W)\bigr)$, each of which lies in $[0,1]$ when $\lambda,\sigma_0(\tau),u^{(t)}_{\tau}(W)\in[0,1]$.
Equation~\eqref{eq:dialectical_update} updates $\mathbf{S}^{(t+1)}$ by elementwise multiplication of $\mathbf{S}^{(t)}$ with $\mathbf{A}(\mathbf{S}^{(t)})$, so every entry remains within $[0,1]$.
\end{proof}

\begin{theorem}[Monotonicity]
\label{thm:mono}
For every unordered pair $\{X,Y\}$ and every $t\ge 0$, we have $S^{(t+1)}_{XY}\le S^{(t)}_{XY}$.
\end{theorem}
\begin{proof}
Each entry of $\mathbf{A}(\mathbf{S}^{(t)})$ lies in $[0,1]$, hence Eq.~\eqref{eq:dialectical_update} gives
$S^{(t+1)}_{XY}=S^{(t)}_{XY}\cdot A_{XY}(\mathbf{S}^{(t)})\le S^{(t)}_{XY}$.
\end{proof}

\begin{theorem}[Pointwise convergence]
\label{thm:conv}
For every unordered pair $\{X,Y\}$, the sequence $\{S^{(t)}_{XY}\}_{t\ge 0}$ converges to a limit $S^{(\ast)}_{XY}\in[0,1]$.
\end{theorem}
\begin{proof}
By Theorem~\ref{thm:bounded} and Theorem~\ref{thm:mono}, each scalar sequence $\{S^{(t)}_{XY}\}_{t\ge 0}$ is bounded below by $0$ and monotone non-increasing, hence it converges in $\mathbb{R}$.
Collecting the limits entrywise yields $\mathbf{S}^{(\ast)}\in[0,1]^{|\mathcal{V}|\times|\mathcal{V}|}$.
\end{proof}

\begin{proposition}[Fixed-point characterization]
Let \(F:[0,1]^{|V|\times |V|}\to[0,1]^{|V|\times |V|}\) be defined by
$F(S)=S\odot A(S)$.
If witness propagation is evaluated over the fixed set of length-2 witness triples in the candidate graph, with attenuation factor
$1-\frac{\lambda}{2}\sigma_0(\tau_{XY})S_{XW}S_{WY}$,
then each entry of \(A(S)\) is a finite product of continuous factors in entries of \(S\). 
Hence \(F\) is continuous on \([0,1]^{|V|\times |V|}\). 
By Theorem~3, \(S^{(t)}\to S^{(*)}\) entrywise, and therefore
\[
S^{(*)}
=
\lim_{t\to\infty}S^{(t+1)}
=
\lim_{t\to\infty}F(S^{(t)})
=
F(S^{(*)}).
\]
Thus the limit state \(S^{(*)}\) is a fixed point of the QACD revision operator.
\end{proposition}

\paragraph{Stability of the Iterative Revision.}
By Theorem~3 and Proposition~1, the infinite revision process converges pointwise to a fixed point of the QACD revision operator under the chosen initialization.
In practice, QACD uses a finite iteration budget \(T_{\max}\) and returns the last iterate once the stopping criterion is met.
Thus, the output should be understood as a finite-horizon stable revision of edge acceptability under the proposed quantitative argumentation semantics, rather than as a claim that all evidential conflicts are resolved by a non-trivial equilibrium.

Unlike greedy constraint-based deletion, this iterative semantics allows the influence of each CI statement to be continuously re-evaluated as the surrounding graph structure evolves, enabling non-greedy conflict resolution under inconsistent statistical evidence.

\textbf{Complexity.}
Let \(n=|V|\) denote the number of variables and let $K$ be the maximum conditioning-set size used in Phase~I. 
Phase~I performs CI tests for each unordered pair $(X,Y)$ over conditioning sets of size at most $K$, yielding a worst-case cost of $O\!\left(n^2 \sum_{k=0}^{K} \binom{n}{k}\right)$ CI tests, as in standard constraint-based pipelines. 
In addition, Phase~I stores at most one representative independence statement $\tau_{XY}$ per variable pair, resulting in a CI-statement pool $\mathcal{T}$ of size $O(n^2)$.

Phase~II iterates a fixed number $T_{\max}$ of dialectical updates. 
For each $\tau_{XY}\in T$, eligible witnesses are common neighbors of $X$ and $Y$, at most $\min\{\deg(X),\deg(Y)\}$ in $G_{\mathrm{cand}}$.
With adjacency-list intersection, each iteration costs $O\!\left(\sum_{X<Y}\min\{\deg(X),\deg(Y)\}\right)$, bounded by \(O(n^3)\) in the worst case and close to \(O(n^2\bar d)\) in sparse graphs, where \(\bar d\) is the average degree of \(G_{\mathrm{cand}}\).
Thus, Phase II adds only a controlled polynomial-time aggregation step.

\subsection{Graph Recovery}
\label{subsec:Graph Recovery}
After $T_{\max}$ iterations (or convergence), we threshold the acceptability matrix to produce a skeleton:
\begin{equation}
X{-}Y \in \hat{G}_{\text{skel}} \quad \Longleftrightarrow \quad S^{(T_{\max})}_{XY}\geq \delta_{0}.
\label{eq:skeleton_threshold}
\end{equation}
This step converts the graded acceptability semantics into a crisp adjacency decision, separating highly supported edges from those dialectically rejected.

We then orient edges by identifying unshielded colliders using the separating sets obtained during Phase~I CI testing. The resulting partially oriented graph is closed under the complete set of Meek orientation rules~\cite{meek1995causal}, yielding the estimated CPDAG $\hat{G}^\ast$.
This recovery procedure mirrors standard constraint-based pipelines, ensuring comparability with classical methods.

\section{Experiments}
\label{sec:experiments}
From an argumentation perspective, the experiments evaluate whether dialectical aggregation of CI evidence improves structural coherence under finite-sample uncertainty. 
In particular, we examine whether treating CI outcomes as defeasible arguments mitigates error propagation in skeleton discovery and leads to more causally valid graphs.

We evaluate QACD on a suite of benchmark Bayesian networks to assess (i) skeleton recovery accuracy and (ii) downstream causal validity.
We compare against representative constraint-based, hybrid, and argumentation-based baselines.

\subsection{Experimental Setup}
We evaluate QACD on standard benchmark networks and compare it against representative causal discovery methods under a unified experimental protocol.

\textbf{Datasets.}
We use eight standard Bayesian networks categorized by size. Statistics are summarized in Table~\ref{tab:dataset_stats}.\footnote{https://www.bnlearn.com/bnrepository/.}
\begin{table}[h]
\centering
\small
\renewcommand{\arraystretch}{1}
\setlength{\tabcolsep}{8pt}
\caption{Benchmark dataset statistics.
$|V|$, $|E|$, and $P$ denote the numbers of nodes, edges, and parameters, respectively.}
\label{tab:dataset_stats}
\begin{tabular}{llccc}
\toprule
Category & Dataset & $|\mathcal{V}|$ & $|\mathcal{E}|$ & $P$\\ 
\midrule
\multirow{3}{*}{\textbf{Small}}
& EARTHQUAKE & 5  & 4  & 10\\
& SURVEY     & 6  & 6  & 21\\
& ASIA       & 8  & 8  & 18\\
\midrule
\multirow{3}{*}{\textbf{Medium}}
& CHILD      & 20 & 25 & 230\\
& INSURANCE  & 27 & 52 & 1008\\
& WATER      & 32 & 66 & 10083\\
\midrule
\multirow{2}{*}{\textbf{Large}}
& HAILFINDER    & 56 & 66 & 2656\\
& WIN95PTS      & 76 & 112 & 574\\
\bottomrule
\end{tabular}
\end{table}

\textbf{Baselines.}
We compare QACD against representative methods from three families:
\textbf{constraint-based} (PC-stable~\cite{spirtes1991algorithm} and MPC~\cite{colombo2014order}),
\textbf{hybrid} (MMHC~\cite{tsamardinos2006max}),
and \textbf{argumentation-based} (ABAPC~\cite{russoargumentative}).
ABAPC is evaluated only on small networks due to runtime constraints.
For the WATER dataset, MMHC is omitted due to deterministic relationships in the data.

\textbf{Protocol.}
All benchmarks are discrete Bayesian networks.
CI testing uses the Pearson $\chi^2$ test with significance level $\alpha=0.05$.
For each network, we sample $N=5000$ i.i.d.\ observations from the ground-truth model and repeat experiments over $50$ random seeds, reporting Mean $\pm$ Std.
All methods are evaluated under the same underlying CI-testing protocol and significance level. The difference is that classical constraint-based baselines use thresholded CI decisions, whereas QACD additionally exploits the full p-value output as graded defeasible evidence; this is part of the modeling design of QACD rather than a difference in the underlying statistical test.

\textbf{QACD Configuration.}
Unless stated otherwise, QACD uses fixed hyperparameters across datasets:
conditioning-set bound $K=3$, $\delta_{\text{cand}}=0.05, \delta_0=0.05$,
$\lambda=0.5$, and $T_{\max}=20$ dialectical iterations.
These hyperparameters are fixed across all datasets and are not tuned separately for individual networks.

\textbf{Metrics.}
We report (1) Skeleton $F_1$ score, (2) Normalized SHD (NSHD), and
(3) Normalized SID (NSID)~\cite{peters2015structural}.
NSHD and NSID are normalized by the number of true edges and therefore need not lie in $[0,1]$.
Lower NSHD and NSID indicate better causal validity.

More details are provided in the supplementary material.

\begin{table*}[t]
\centering
\small
\renewcommand{\arraystretch}{1}
\setlength{\tabcolsep}{8pt}
\caption{\textbf{Small Networks Results.} Comparison of Skeleton $F_1$, Normalized SHD (NSHD), and Normalized SID (NSID). All values are reported as \textbf{Mean $\pm$ Std}. \textbf{Bold} indicates the best performance (higher is better for $F_1$, lower is better for NSHD/NSID).}
\label{tab:small_main}
\begin{tabular}{llrrrr}
\toprule

Dataset & Method & $F_1$ $\uparrow$ & NSHD $\downarrow$ & NSID$_{\mathrm{low}}$ $\downarrow$ & NSID$_{\mathrm{high}}$ $\downarrow$ \\
\midrule
\multirow{5}{*}{\textbf{ASIA}}
& PC     & $0.8724 \pm 0.0341$ & $0.5750 \pm 0.1118$ & $1.5500 \pm 0.7984$ & $2.4250 \pm 0.7984$ \\
& MPC & $\mathbf{0.9340 \pm 0.0344}$ & $0.3875 \pm 0.0379$ & $0.7875 \pm 0.7939$ & $1.4050 \pm 0.3733$ \\
& MMHC   & $0.7487 \pm 0.0459$ & $0.7500 \pm 0.0000$ & $1.8250 \pm 0.1677$ & $3.0750 \pm 0.1677$ \\
& ABAPC  & $0.5000 \pm 0.1800$ & $0.6550 \pm 0.2825$ & $1.4650 \pm 0.8488$ & $4.1900 \pm 0.9900$ \\
& \textbf{QACD(ours)} & $0.9181 \pm 0.0341$ & $\mathbf{0.3750 \pm 0.0000}$ & $\mathbf{0.2250 \pm 0.1630}$ & $\mathbf{1.1000 \pm 0.1630}$ \\
\midrule
\multirow{5}{*}{\textbf{EARTHQUAKE}}
& PC     & $\mathbf{1.0000 \pm 0.0000}$ & $\mathbf{0.0000 \pm 0.0000}$ & $\mathbf{0.0000 \pm 0.0000}$ & $\mathbf{0.0000 \pm 0.0000}$ \\
& MPC & $0.9978 \pm 0.0157$ & $0.0050 \pm 0.0354$ & $\mathbf{0.0000 \pm 0.0000}$ & $\mathbf{0.0000 \pm 0.0000}$ \\
& MMHC   & $\mathbf{1.0000 \pm 0.0000}$ & $\mathbf{0.0000 \pm 0.0000}$ & $\mathbf{0.0000 \pm 0.0000}$ & $\mathbf{0.0000 \pm 0.0000}$ \\
& ABAPC  & $0.9400 \pm 0.0600$ & $0.1400 \pm 0.1450$ & $\mathbf{0.0000 \pm 0.0000}$ & $2.3300 \pm 1.1350$ \\
& \textbf{QACD(ours)} & $\mathbf{1.0000 \pm 0.0000}$ & $\mathbf{0.0000 \pm 0.0000}$ & $\mathbf{0.0000 \pm 0.0000}$ & $\mathbf{0.0000 \pm 0.0000}$ \\
\midrule
\multirow{5}{*}{\textbf{SURVEY}}
& PC     & $0.6489 \pm 0.1280$ & $0.5667 \pm 0.1491$ & $2.3000 \pm 0.0745$ & $2.3000 \pm 0.0745$ \\
& MPC & $0.6196 \pm 0.1491$ & $0.6167 \pm 0.1438$ & $2.4500 \pm 0.2189$ & $2.4833 \pm 0.2901$ \\
& MMHC   & $\mathbf{0.8218 \pm 0.0488}$ & $0.9000 \pm 0.2236$ & $\mathbf{1.3000 \pm 0.4314}$ & $3.2000 \pm 0.9531$ \\
& ABAPC  & $0.4200 \pm 0.1100$ & $0.8030 \pm 0.1830$ & $2.3700 \pm 0.6080$ & $2.7360 \pm 0.4610$ \\
& \textbf{QACD(ours)} & $0.7588 \pm 0.0604$ & $\mathbf{0.4333 \pm 0.1491}$ & $1.9333 \pm 0.3651$ & $\mathbf{1.9333 \pm 0.3651}$ \\
\bottomrule
\end{tabular}
\end{table*}

\begin{table*}[t]
\centering
\small
\renewcommand{\arraystretch}{1}
\setlength{\tabcolsep}{8pt}
\caption{\textbf{Medium and Large Networks Results.}}
\label{tab:large_networks}
\begin{tabular}{llrrrr}
\toprule
Dataset & Method & $F_1$ $\uparrow$ & NSHD $\downarrow$ & NSID$_{\mathrm{low}}$ $\downarrow$ & NSID$_{\mathrm{high}}$ $\downarrow$ \\
\midrule
\multirow{4}{*}{\textbf{INSURANCE}}
& PC     & $0.6664 \pm 0.0300$ & $0.6846 \pm 0.0617$ & $10.8000 \pm 0.7295$ & $10.8154 \pm 0.7054$ \\
& MPC & $0.6748 \pm 0.0208$ & $0.6500 \pm 0.0211$ & $10.3769 \pm 0.4361$ & $10.3885 \pm 0.4496$ \\
& MMHC   & $0.6335 \pm 0.0265$ & $0.8000 \pm 0.0258$ & $9.5923 \pm 0.2073$ & $10.6923 \pm 0.2159$ \\
& \textbf{QACD(ours)} & $\mathbf{0.7361 \pm 0.0301}$ & $\mathbf{0.5692 \pm 0.0554}$ & $\mathbf{8.7462 \pm 0.4713}$ & $\mathbf{8.8577 \pm 0.5437}$ \\
\midrule
\multirow{4}{*}{\textbf{CHILD}}
& PC     & $0.8445 \pm 0.0449$ & $0.6400 \pm 0.1020$ & $8.3040 \pm 1.0196$ & $10.5440 \pm 0.9926$ \\
& MPC & $\mathbf{0.9008 \pm 0.0263}$ & $0.5440 \pm 0.0607$ & $7.2080 \pm 0.2252$ & $9.1520 \pm 0.7124$ \\
& MMHC   & $0.8425 \pm 0.0118$ & $0.8400 \pm 0.0000$ & $\mathbf{4.2160 \pm 0.3220}$ & $12.4560 \pm 0.0358$ \\
& \textbf{QACD(ours)} & $0.8861 \pm 0.0232$ & $\mathbf{0.4400 \pm 0.1020}$ & $4.2400 \pm 0.9104$ & $\mathbf{5.4320 \pm 1.8156}$ \\
\midrule
\multirow{3}{*}{\textbf{WATER}}
& PC & $0.4550 \pm 0.0244$ & $0.9000 \pm 0.0275$ & $7.1303 \pm 0.5925$ & $\mathbf{8.1545 \pm 0.2966}$ \\
& MPC & $0.4898 \pm 0.0201$ & $\mathbf{0.8758 \pm 0.0127}$ & $6.7091 \pm 0.5812$ & $8.2030 \pm 0.3152$ \\
& \textbf{QACD(ours)} & $\mathbf{0.4931 \pm 0.0094}$ & $0.9152 \pm 0.0380$ & $\mathbf{6.1242 \pm 0.2905}$ & $8.2818 \pm 0.5038$ \\
\midrule
\multirow{4}{*}{\textbf{HAILFINDER}}
& PC     & $0.5489 \pm 0.0131$ & $1.3788 \pm 0.0283$ & $9.1455 \pm 0.6605$ & $10.1000 \pm 0.4164$ \\
& MPC & $0.5536 \pm 0.0107$ & $1.4061 \pm 0.0392$ & $9.8788 \pm 0.6359$ & $10.8970 \pm 0.6722$ \\
& MMHC   & $0.6490 \pm 0.0091$ & $\mathbf{0.6576 \pm 0.0136}$ & $9.8455 \pm 0.0166$ & $13.4061 \pm 0.0166$ \\
& \textbf{QACD(ours)} & $\mathbf{0.7193 \pm 0.0293}$ & $0.6788 \pm 0.0984$ & $\mathbf{7.0545 \pm 1.3054}$ & $\mathbf{9.6333 \pm 2.4174}$ \\
\midrule
\multirow{4}{*}{\textbf{WIN95PTS}}
& PC     & $0.6668 \pm 0.0367$ & $0.6589 \pm 0.0487$ & $4.8964 \pm 0.6593$ & $\mathbf{5.7964 \pm 0.5919}$ \\
& MPC & $0.6353 \pm 0.0311$ & $0.7000 \pm 0.0436$ & $5.4089 \pm 0.4750$ & $6.3339 \pm 0.3662$ \\
& MMHC   & $\mathbf{0.7147 \pm 0.0136}$ & $\mathbf{0.6339 \pm 0.0089}$ & $6.0446 \pm 0.4247$ & $7.8006 \pm 0.2481$ \\
& \textbf{QACD(ours)} & $0.6870 \pm 0.0321$ & $0.6429 \pm 0.0296$ & $\mathbf{4.7464 \pm 0.5152}$ & $6.4054 \pm 0.3719$ \\
\bottomrule
\end{tabular}
\end{table*}

\subsection{Structural Accuracy and Causal Validity Across Benchmarks}

Tables~\ref{tab:small_main} and~\ref{tab:large_networks} summarize results across small, medium, and large networks. 
Overall, QACD consistently produces structures with strong causal validity (low NSHD/NSID) while maintaining competitive skeleton accuracy across datasets of varying complexity.

QACD shows the greatest efficacy on networks where CI evidence is noisy or inconsistent. 
On INSURANCE, QACD achieves the highest $F_1$ and the lowest NSHD while substantially tightening NSID bounds relative to PC, MPC, and MMHC. 
Similar behavior appears on SURVEY and HAILFINDER, where QACD improves NSHD and consistently reduces NSID$_{\mathrm{high}}$. 

Across several datasets (e.g., ASIA and CHILD), QACD exhibits a recurring trade-off between recall and structural coherence. 
While aggressive baselines such as MPC or MMHC sometimes achieve slightly higher skeleton $F_1$, QACD produces lower NSHD and tighter NSID bounds, indicating superior interventional reliability. 

On datasets with deterministic relationships (e.g., WATER), QACD remains stable and improves the lower SID estimate while remaining comparable on the upper SID estimate, even when greedy methods become sensitive to statistical degeneracy.
Because argument strengths are attenuated rather than globally optimized, the method avoids instability caused by degenerate sufficient statistics.

On low-complexity networks (e.g., EARTHQUAKE), PC, MMHC, and QACD all recover the ground-truth structure, confirming that dialectical aggregation maintains robustness when CI evidence is already consistent. On WIN95PTS, QACD remains competitive but does not dominate, reflecting the expected trade-off of conservative filtering in regimes where CI evidence is highly reliable.

Overall, the results indicate that QACD is most beneficial when CI evidence is
noisy or inconsistent. By revising edge acceptability through connectivity-mediated propagation, dialectical aggregation improves global structural coherence and reduces interventional error, while remaining competitive in skeleton recovery across datasets of varying sizes.

\subsection{Ablation: Effect of Dialectical Aggregation}

\begin{table}[ht]
\centering
\small
\renewcommand{\arraystretch}{1.1}
\setlength{\tabcolsep}{5pt}
\caption{\textbf{Ablation Study.} Comparison of the full dialectical aggregation model ($T_{\max}=20$) against the initial candidate skeleton before aggregation ($T_{\max}=0$).}
\label{tab:ablation}
\begin{tabular}{l|cc|cc}
\toprule
& \multicolumn{2}{c|}{\textbf{QACD ($T_{\max}=20$)}} & \multicolumn{2}{c}{\textbf{Ablation ($T_{\max}=0$)}} \\
Dataset & $F_1$ & NSID$_{\mathrm{low}}$ & $F_1$ & NSID$_{\mathrm{low}}$ \\
\midrule
ASIA       & $\mathbf{0.9181}$ & $\mathbf{0.2250}$ & $0.7733$ & $2.2500$ \\
EARTHQUAKE & $\mathbf{1.0000}$ & $\mathbf{0.0000}$ & $0.8756$ & $\mathbf{0.0000}$ \\
SURVEY     & $\mathbf{0.7588}$ & $\mathbf{1.9333}$ & $0.5127$ & $2.4667$ \\
\bottomrule
\end{tabular}
\end{table}

To isolate the impact of the dialectical aggregation phase, we compare full QACD with the initial candidate skeleton before aggregation, i.e., the variant with $T_{\max}=0$, as shown in Table~\ref{tab:ablation}.
Across datasets, removing the iterative aggregation phase substantially degrades both skeleton quality and causal validity.
The effect is most pronounced on networks where short-range witness connectivity provides reliable structural cues, confirming the role of dialectical aggregation in stabilizing skeleton discovery under noisy CI evidence.

\subsection{Impact of Sample Size}
We analyze the sensitivity of QACD and PC to sample size on the INSURANCE network (\(N\in\{500,1000,2000,5000\}\); Figure~\ref{fig:sample_size}), reporting Mean \(\pm\) Std over 5 runs. Using PC as the canonical constraint-based baseline isolates the contribution of dialectical aggregation under identical CI inputs. QACD outperforms PC at \(N=500\) (\(0.5443\pm0.0172\) vs.\ \(0.5114\pm0.0520\)), \(N=1000\) (\(0.6383\pm0.0275\) vs.\ \(0.5414\pm0.0536\)), and \(N=2000\) (\(0.6968\pm0.0388\) vs.\ \(0.6420\pm0.0362\)); at \(N=5000\), the gap narrows (\(0.7361\pm0.0301\) vs.\ \(0.7267\pm0.0362\)). This suggests that dialectical aggregation is most helpful in low-sample regimes and becomes less advantageous as CI decisions stabilize with increasing \(N\).

\begin{figure}[t]
    \centering
    \includegraphics[width=1\linewidth]{insurance_sample_size.pdf}
    \caption{\textbf{Sample Size Efficiency on INSURANCE.} QACD (orange) outperforms PC (blue) in finite-sample regimes and maintains competitive asymptotic performance as $N$ increases.}
    \label{fig:sample_size}
\end{figure}

QACD is intentionally conservative: dialectical aggregation suppresses weakly supported edges, which can reduce recall in dense graphs.
In settings where CI evidence is already highly consistent or sample sizes are very large, the advantage of non-monotonic conflict resolution diminishes.
These cases reflect a design trade-off rather than instability of the proposed method.

\subsection{Discussion} 
The preceding results reveal a consistent pattern rather than a uniform dominance claim.
The results suggest that QACD is most beneficial when CI evidence is noisy or mutually inconsistent. On datasets such as INSURANCE, SURVEY, and HAILFINDER, dialectical aggregation improves NSHD and NSID while maintaining competitive skeleton \(F_1\). This is consistent with treating CI outcomes as defeasible arguments: weak or conflicting CI claims are attenuated through connectivity-mediated revision, reducing the propagation of local CI errors into global structural mistakes. As sample size increases and CI decisions become more stable, the benefit of dialectical conflict resolution correspondingly diminishes. 

Across datasets such as ASIA and CHILD, QACD exhibits a recurring trade-off between skeleton recall and causal validity. While aggressive baselines such as MPC or MMHC sometimes achieve slightly higher \(F_1\), QACD often yields lower NSHD and tighter NSID bounds, indicating improved structural coherence and interventional reliability. This reflects its conservative revision strategy: rather than making irreversible early commitments, QACD iteratively suppresses weakly supported adjacencies. 

On datasets with deterministic relations, such as WATER, QACD remains stable and improves some intervention-oriented metrics even when greedy or likelihood-based methods become sensitive to degeneracy; conversely, when CI evidence is already highly consistent or informative, as in EARTHQUAKE or WIN95PTS, dialectical aggregation provides more limited additional benefit.

Overall, the experiments support the usefulness of viewing finite-sample causal discovery as reasoning over conflicting statistical arguments. 
In the evaluated noisy or inconsistent CI regimes, dialectical aggregation often improves global coherence and reduces interventional error while remaining competitive in skeleton recovery.

\section{Related Work}
\label{sec:related_work}

This work relates to both causal discovery and computational argumentation, especially methods for handling uncertainty and conflict in structure learning.

\subsection{Causal Discovery from Observational Data}

\textbf{Constraint-based approaches}, exemplified by the PC algorithm~\cite{spirtes2000causation} and Fast Causal Inference~\cite{spirtes1995causal}, infer causal structure by testing CI relations and compiling accepted independences into graph edits. 
Under the Markov and Faithfulness assumptions, these methods are asymptotically consistent, but in finite-sample regimes they are sensitive to CI errors: a single false independence or dependence may trigger incorrect edge deletions and propagate to later orientation steps ~\cite{uhler2013geometry}. 
Variants such as PC-stable reduce order dependence~\cite{colombo2014order}, and multiple-testing corrections can improve statistical robustness, but most constraint-based pipelines still rely on hard CI decisions that become irreversible once incorporated into the graph. QACD instead delays irreversible commitments by aggregating CI outcomes as graded defeasible evidence through an iterative acceptability semantics.

\textbf{Score-based methods}, such as Greedy Equivalence Search and its extensions~\cite{chickering2002optimal,ramsey2017million}, formulate causal discovery as search over graph structures using penalized likelihood criteria. 
Recent continuous optimization approaches, including NOTEARS and subsequent variants ~\cite{zheng2018dags,ng2024structure}, further cast structure learning as differentiable optimization with acyclicity constraints. 
These methods provide important alternatives to purely constraint-based discovery, but they do not explicitly model conflicts among CI outcomes as argumentative interactions.
QACD occupies a complementary position: it preserves the local efficiency of CI testing while introducing a quantitative argumentation layer for resolving evidential conflicts during skeleton discovery.

\subsection{Argumentation in Artificial Intelligence}

\textbf{Abstract and quantitative argumentation.}
Since Dung’s seminal formulation of AFs~\cite{dung1995acceptability}, argumentation has become a foundational paradigm for non-monotonic reasoning under inconsistency~\cite{li2025investigation,liao2024attack}. QAFs extend this setting by assigning intrinsic strengths to arguments and defining semantics based on graded acceptability~\cite{chi2022quantitative,rago2016discontinuity}. Unlike ~\cite{amgoud2018weighted}, which models arguments in a bipolar framework, QACD treats conditional independence tests as defeasible evidence. This distinction allows QACD to better handle noisy or uncertain data, providing a more robust framework for causal discovery.

\textbf{Argumentation for causal inquiry.}
The use of argumentation in causal reasoning is a growing but still relatively specialized area.
Early work primarily focused on explanation, employing argumentative structures to justify or communicate causal claims to users~\cite{vcyras2021argumentative}. 
More recent studies have explored argumentation-based mechanisms for causal discovery and conflict handling~\cite{russoargumentative,li2026enhancing}. 
Related probabilistic approaches connect uncertainty over arguments with graded acceptability ~\cite{hunter2017probabilistic}.

The work most closely related to ours is ABAPC~\cite{russoargumentative}, which augments PC with argumentation mechanisms mainly for resolving orientation-level conflicts such as competing v-structures.
In contrast, QACD integrates quantitative argumentation directly into skeleton discovery.
By treating CI statements as graded attackers of adjacency hypotheses and propagating their influence through witness-based structural context, QACD enables non-monotonic revision of edge hypotheses before the orientation phase.

\section{Conclusion}
\label{sec:conclusion}

We presented QACD, a QAF for causal discovery that treats CI outcomes as defeasible evidence and aggregates them under structural context.
Rather than compiling finite-sample CI decisions into irreversible constraints, QACD revises edge acceptability through direct CI attacks and witness-based propagated attenuation.
This provides an iterative, semantics-driven mechanism for resolving conflicting statistical evidence during skeleton discovery.

Empirically, QACD yields more coherent structures and improved interventional validity in several regimes where CI evidence is noisy or inconsistent, while remaining competitive in skeleton recovery.
These results support the view that finite-sample causal discovery can be fruitfully formulated as reasoning over graded and potentially conflicting statistical arguments.

Several directions remain open.
First, while this paper focuses on the causally sufficient setting, the framework could be extended to latent confounding by introducing arguments for bi-directed edges in ancestral graphs.
Second, the iterative update could be accelerated for very large graphs using approximate message-passing or sparsity-aware implementations.
Third, although we expect finite-sample conflicts to diminish under standard Markov and faithfulness assumptions with consistent CI testing, a complete oracle-level consistency analysis for QACD remains future work.
Finally, interventional data could be incorporated as a new class of intervention arguments, enabling argumentation-based discovery from mixed observational and experimental evidence.

\bibliographystyle{KR26_authors_kit/kr}
\bibliography{KR26_authors_kit/kr}

\clearpage
\appendix

\section{Orientation Rules (Meek Rules)}
\label{app:meek_rules}

After the dialectical aggregation phase determines the final skeleton and identifies unshielded colliders (v-structures), QACD applies the standard Meek orientation rules \cite{meek1995causal} to orient the remaining undirected edges. These rules are applied iteratively to closure, while avoiding the introduction of new unshielded colliders or directed cycles, yielding a CPDAG representation of the corresponding Markov equivalence class.

The four rules are applied iteratively until no further edges can be oriented:
\begin{itemize}
    \item \textbf{R1 (Chain Propagation):} If there is a directed edge $X \to Y$ and an undirected edge $Y - Z$ such that $X$ and $Z$ are not adjacent, then orient $Y - Z$ as $Y \to Z$. This avoids creating a new unshielded collider at $Y$.
    
    \item \textbf{R2 (Cycle Avoidance):} If there is a directed path $X \to Z \to Y$ and an undirected edge $X - Y$, then orient $X - Y$ as $X \to Y$ to avoid creating a directed cycle.
    
    \item \textbf{R3 (Double-Triangle Rule):} If there are undirected edges $X - Z$, $X - W$, and $X - Y$, and directed edges $Z \to Y$ and $W \to Y$, where $Z$ and $W$ are not adjacent, then orient $X - Y$ as $X \to Y$.
    
    \item \textbf{R4 (Triangle-Chain Rule):} If there are undirected edges $X - Z$ and $X - Y$, and directed edges $Z \to W$ and $W \to Y$, where $Z$ and $Y$ are not adjacent, then orient $X - Y$ as $X \to Y$.
\end{itemize}

\section{Evaluation Metrics}
\label{app:metrics}

We evaluate the performance of QACD using three standard metrics that capture different aspects of structural and causal accuracy. Let $G^\ast$ be the ground-truth graph and $\hat{G}$ be the estimated CPDAG.

\textbf{Skeleton $F_1$ Score.}
This metric evaluates the accuracy of the undirected adjacency structure (skeleton). It is the harmonic mean of precision and recall:
\[
F_1 = \frac{2 \cdot \text{Precision} \cdot \text{Recall}}
{\text{Precision} + \text{Recall}},
\]
where Precision is the fraction of estimated skeleton edges that are present in $G^\ast$, and Recall is the fraction of true skeleton edges correctly identified by $\hat{G}$.

\textbf{Normalized Structural Hamming Distance (NSHD).}
The SHD measures the total number of edge modifications, including additions, deletions, or reversals, required to transform $\hat{G}$ into $G^\ast$. To facilitate comparison across networks of different sizes, we report the normalized SHD:
\[
\text{NSHD} =
\frac{\text{SHD}(\hat{G}, G^\ast)}
{|E^\ast|},
\]
where $|E^\ast|$ denotes the number of edges in the ground-truth graph. Lower NSHD indicates better overall structural recovery.

\textbf{Normalized Structural Intervention Distance (NSID).}
Unlike SHD, which treats all edge errors equally, the SID \cite{peters2015structural} assesses the causal validity of the graph by counting the number of incorrectly predicted interventional distributions between pairs of variables. Since $\hat{G}$ is a CPDAG, we report lower and upper bounds over the corresponding Markov equivalence class. The normalized version is computed as
\[
\text{NSID} =
\frac{\text{SID}(\hat{G}, G^\ast)}
{|E^\ast|}.
\]
We report $\text{NSID}_{\text{low}}$ and $\text{NSID}_{\text{high}}$ to represent the lower and upper bounds of the distance over the estimated equivalence class. As with NSHD, lower values indicate better causal validity.

\subsection{Metric Computation}
Skeleton precision, recall, and $F_1$ are computed in Python by comparing the
undirected skeletons of the estimated graph and the ground-truth graph.
For SHD and SID, we use the R packages \texttt{pcalg} and \texttt{SID} through
\texttt{rpy2}. In particular, SHD is computed using \texttt{pcalg::shd}, and
SID bounds are computed using \texttt{SID::structIntervDist}. Since the
estimated graph is a CPDAG, we report lower and upper SID bounds over the
corresponding Markov equivalence class.

\section{Implementation Details}

\subsection{Computational Environment}
Experiments were conducted on a dual-socket Intel Xeon Gold 6330 server
with 372 GB RAM running Ubuntu Linux. All experiments were implemented in
Python 3.10.18 and executed on CPU only.

\subsection{CPDAG Representation}
In the implementation, CPDAGs are represented using directed NetworkX graphs.
A directed edge $X \to Y$ is represented by a single directed arc, whereas an
undirected edge $X-Y$ is represented by two opposite arcs, $X \to Y$ and
$Y \to X$. This representation is used consistently for QACD, the baselines,
and metric computation.

\subsection{CI-Statement Pool Construction}
In Phase~I of QACD, the conditioning-set size is bounded by $K=3$.
For each unordered variable pair $(X,Y)$, the unconditional CI test is always
performed. For each conditioning-set size $k \in \{1,\ldots,K\}$, we evaluate
at most $50$ candidate conditioning sets for computational efficiency.

Among the tested conditioning sets, the independence statement with the largest
$p$-value is retained as the representative CI argument for $(X,Y)$. This yields
a CI-statement pool $\mathcal{T}$ with size $O(|\mathcal{V}|^2)$, which is used
in Phase~II dialectical aggregation.

\subsection{Metric Computation}
Skeleton precision, recall, and $F_1$ are computed in Python by comparing the
undirected skeletons of the estimated graph and the ground-truth graph. For SHD
and SID, we use the R packages \texttt{pcalg} and \texttt{SID} through
\texttt{rpy2}. SHD is computed using \texttt{pcalg::shd}, and SID bounds are
computed using \texttt{SID::structIntervDist}. Since the estimated graph is a
CPDAG, we report lower and upper SID bounds over the corresponding Markov
equivalence class.

\subsection{Hyperparameter Settings}
Unless otherwise stated, QACD uses the same hyperparameter settings across all
datasets: CI significance level $\alpha=0.05$, candidate threshold
$\delta_{\mathrm{cand}}=0.05$, final acceptability threshold $\delta_0=0.05$,
maximum conditioning-set size $K=3$, attack decay $\lambda=0.5$, maximum
dialectical iterations $T_{\max}=20$, and convergence tolerance $10^{-4}$.

\end{document}